\documentclass{article}
\usepackage{microtype}
\usepackage{graphicx}
\usepackage{url}
\usepackage{xcolor}
\usepackage{multirow}
\usepackage{graphicx}
\usepackage{adjustbox}
\usepackage{subcaption}
\usepackage{booktabs} % for professional tables
\usepackage{caption}

\definecolor{baselinecolor}{gray}{0.95}
\newcommand{\baseline}[1]{\colorbox{baselinecolor}{#1}}
\definecolor{bestsettingcolor}{RGB}{255,255,180}  % This defines a yellow color
\newcommand{\bestsetting}[1]{\colorbox{bestsettingcolor}{#1}}
\usepackage{hyperref}
\usepackage[accepted]{icml2025}
\usepackage{amsmath}
\usepackage{amssymb}
\usepackage{mathtools}
\usepackage{amsthm}
\usepackage[capitalize,noabbrev]{cleveref}
\theoremstyle{plain}

\theoremstyle{definition}

\theoremstyle{remark}

\icmltitlerunning{AirCast}
\begin{document}
\twocolumn[
% \icmltitle{AirCast: Improving Air Pollution Forecasting \\ Through A Multi-Variable Approach}
\icmltitle{AirCast: Improving Air Pollution Forecasting \\ Through Multi-Variable Data Alignment}
% It is OKAY to include author information, even for blind
% submissions: the style file will automatically remove it for you
% unless you've provided the [accepted] option to the icml2025
% package.
% List of affiliations: The first argument should be a (short)
% identifier you will use later to specify author affiliations
% Academic affiliations should list Department, University, City, Region, Country
% Industry affiliations should list Company, City, Region, Country
% You can specify symbols, otherwise they are numbered in order.
% Ideally, you should not use this facility. Affiliations will be numbered
% in order of appearance and this is the preferred way.
\icmlsetsymbol{equal}{*}
\begin{icmlauthorlist}
\icmlauthor{Vishal Nedungadi}{wur,mbzuai}
\icmlauthor{Muhammad Akhtar Munir}{mbzuai}
\icmlauthor{Marc Rußwurm}{wur}
\icmlauthor{Ron Sarafian}{wis}
\icmlauthor{Ioannis N. Athanasiadis}{wur}
\icmlauthor{Yinon Rudich}{wis}
\icmlauthor{Fahad Shahbaz Khan}{mbzuai,lu}
%\icmlauthor{}{sch}
\icmlauthor{Salman Khan}{mbzuai,anu}
%\icmlauthor{}{sch}
%\icmlauthor{}{sch}
\end{icmlauthorlist}
\icmlaffiliation{wur}{Wageningen University and Research}
\icmlaffiliation{mbzuai}{Mohamed bin Zayed University of AI}
\icmlaffiliation{anu}{Australian National University}
\icmlaffiliation{lu}{Linköping University}
\icmlaffiliation{wis}{Weizmann Institute of Science}

\icmlcorrespondingauthor{Vishal Nedungadi}{vishal.nedungadi@wur.nl}
% You may provide any keywords that you
% find helpful for describing your paper; these are used to populate
% the "keywords" metadata in the PDF but will not be shown in the document
% \icmlkeywords{Machine Learning, ICML}
\vskip 0.3in
]

% this must go after the closing bracket ] following \twocolumn[ ...

% This command actually creates the footnote in the first column
% listing the affiliations and the copyright notice.
% The command takes one argument, which is text to display at the start of the footnote.
% The \icmlEqualContribution command is standard text for equal contribution.
% Remove it (just {}) if you do not need this facility.

\printAffiliationsAndNotice{}  % leave blank if no need to mention equal contribution
% \printAffiliationsAndNotice{\icmlEqualContribution} % otherwise use the standard text.

\begin{abstract}
Air pollution remains a leading global health risk, exacerbated by rapid industrialization and urbanization, contributing significantly to morbidity and mortality rates.
% Accurate forecasting of particulate matter (PM1, PM2.5, PM10) is crucial to implement effective mitigation strategies.
% Accurately forecasting particulate matter (PM1, PM2.5, PM10) is essential for developing effective strategies to mitigate health and environmental impacts.
In this paper, we introduce \textit{AirCast}, a novel multi-variable air pollution forecasting model, by combining weather and air quality variables.
\textit{AirCast} employs a multi-task head architecture that simultaneously forecasts atmospheric conditions and pollutant concentrations, improving its understanding of how weather patterns affect air quality. 
% As a case study, we apply this approach to the Middle East and North Africa (MENA) region, demonstrating its effectiveness in capturing regional PM patterns.
% Accurately predicting extreme PM concentrations is challenging due to their rare occurrence in the data. To address this, we introduce a Frequency-weighted Mean Absolute Error (fMAE) loss function, which places greater emphasis on these cases during training.
Predicting extreme pollution events is challenging due to their rare occurrence in historic data, resulting in a heavy-tailed distribution of pollution levels. To address this, we propose a novel Frequency-weighted Mean Absolute Error (fMAE) loss, adapted from the class-balanced loss for regression tasks. Informed from domain knowledge, we investigate the selection of key variables known to influence pollution levels. Additionally, we align existing weather and chemical datasets across spatial and temporal dimensions. \textit{AirCast}'s integrated approach, combining multi-task learning, frequency weighted loss and domain informed variable selection, enables more accurate pollution forecasts.
Our source code and models are made public here (\url{https://github.com/vishalned/AirCast.git})

% Our integrated approach enables more accurate predictions of PM levels.
\end{abstract}

\section{Introduction}
Rapid industrialization, economic growth, and climate change have significantly worsened air pollution \citep{Nakhjiri2024}, raising serious concerns about environmental quality and public health. 
% Accurate forecasting of air pollution is critical, as it allows policymakers to take timely actions to reduce vulnerability and mitigate health risks.
Among various pollutants, particulate matter such as PM1, PM2.5, and PM10 (particles smaller than 1, 2.5, and 10 micrometers, respectively) have been directly associated with adverse health effects.
These tiny particles can penetrate the respiratory system, possibly leading to cancer and various respiratory and cardiovascular diseases.
The World Health Organization (WHO) reports that around 99\% of the global population is exposed to air that does not meet its 2019 quality guidelines. 
According to recent estimates \citep{whoAmbientoutdoor}, air pollution is responsible for approximately 6.7 million premature deaths annually. 
This highlights the urgent need for improved forecasting methods to accurately predict air quality. 
These forecasts can advise policy decisions and contribute to reducing emissions strategies. 
% https://www.who.int/news-room/fact-sheets/detail/ambient-(outdoor)-air-quality-and-health

Air pollution forecasting for PM primarily relies on two approaches: physics-based and data-driven models. 
% Physics-based models employ fundamental atmospheric chemistry and physics principles to simulate the dispersion and chemical transformation of pollutants. 
Physics-based models simulate pollutant dispersion and chemical transformations using fundamental principles of atmospheric chemistry and physics.
These models often use non-linear empirical methods \citep{cobourn2010enhanced, lv2016development} to represent complex environmental interactions.
% While they offer valuable insights into the underlying physical and chemical processes concerning air quality, the physics-based models often meet accuracy limitations. 
% The highly dynamic and complex nature of atmospheric processes makes it challenging for these models to precisely capture both long-term and short-term trends. 
% Although they provide valuable insight into the physical and chemical processes that govern air quality, their accuracy is often limited due to the highly dynamic and complex nature of atmospheric processes, making it challenging to precisely capture both long-term and short-term trends.
Although they offer valuable insight into the physical and chemical processes that govern air quality, their accuracy is often constrained by the dynamic complexity of atmospheric systems. 
This makes it difficult to precisely capture both long and short-term trends.
In contrast, data-driven approaches \citep{bi2023accurate, nguyen2023climaxfoundationmodelweather, nguyen2023scalingtransformerneuralnetworks,bodnar2024aurora} utilize machine learning methodologies to model complex relationships among various atmospheric variables, such as temperature, wind, and PM concentrations. 
These models are trained to capture non-linear patterns and dependencies implicitly under diverse atmospheric conditions. 
Moreover, data-driven models adapt more readily to new data and evolving environmental conditions than physics-based models, identifying patterns and relationships that physics-based models cannot explicitly represent. 
% Existing data-driven approaches are yet to make use of other variables that could potentially impact PM concentrations.
Existing data-driven approaches overlook variables that could potentially influence PM concentrations.

% In this work, we show how the foundational backbones available for predicting weather variables can be used to forecast air quality variables. 
In this work, we enhance PM forecasting methods by integrating weather and air quality variables to improve accuracy. 
Our proposed model, AirCast, is a Vision Transformer (ViT) \citep{dosovitskiy2021imageworth16x16words}, designed for air pollution forecasting by adapting a weather foundational model \cite{nguyen2023climaxfoundationmodelweather}. 
% AirCast leverages the ability of large-scale pre-trained models to learn generalizable representations from diverse datasets, improving performance in downstream tasks like air quality prediction. 
By utilizing large-scale pre-trained models, AirCast learns generalizable representations from diverse datasets, enhancing its performance in air quality prediction.
An important aspect of this adaptation is our development of a combined dataset integrating weather and air quality variables for precise air pollution forecasting.
We source weather variables from WeatherBench \citep{Rasp_2020} and air quality variables from the Copernicus Atmosphere Monitoring Service (CAMS) EAC4 dataset \citep{CAMS23}. 
This multi-variable approach allows AirCast to capture the complex relationships between weather conditions and pollutant levels. 
Similar to \cite{nguyen2023climaxfoundationmodelweather}, the model architecture incorporates variable tokenization and variable aggregation modules to efficiently handle a large number of variables and reduce the sequence length. 
To further enhance its capabilities, a multi-task head architecture enables the model to predict both atmospheric weather and air pollution variables simultaneously. 
Additionally, a Frequency-weighted Mean Absolute Error (fMAE) loss function inspired by the class balanced loss function \citep{cui2019classbalancedlossbasedeffective} addresses the heavy-tailed distributions of pollutants, improving the accuracy of predictions for extreme cases. Furthermore, learning from domain knowledge we also investigate the selection of key variables known to affect PM concentrations.

Our study focuses on the Middle East and North Africa (MENA) region, which consistently experiences some of the highest levels of PM concentrations globally, often exceeding the recommended air quality standards of the WHO \citep{heger2022blue}.
Forecasting air pollution in the MENA region using data-driven methods is extremely important due to its distinct environmental challenges. 
The challenges include frequent dust storms, industrial emissions, reliance on fossil fuels, rapid urbanization, and low rainfall, all of which combine to significantly degrade air quality.
%Our study focuses on predicting particulate matter levels in the MENA region due to: its critical atmospheric conditions and interventions to mitigate the adverse effects of air pollution.
In this paper, we focus on accurate and efficient PM forecasting in MENA region, aiming to support mitigation efforts and reduce the harmful effects of air pollution.
% To this end, our work also demonstrates how the knowledge learned by the global data-driven models can be efficiently adapted to regional conditions. 
%To tailor the model to the MENA region, the model transfers knowledge learned from global weather to regional conditions. 
% In this paper, we address the challenges of air pollution forecasting in the MENA region by presenting AirCast. 
% To improve the model's generalization abilities, randomized lead time is adopted during training. This practice acts as data augmentation, improving the model's performance across different forecasting ranges. 
Our main contributions are as follows:
\begin{enumerate}
    \item \textbf{Integrated Forecasting:}: 
    To capture the interactions between weather and air quality, we develop a multi-task head architecture that simultaneously predicts atmospheric and pollution variables.
    % To enable the model to capture the complex interactions between weather and air quality patterns, we develop a multi-task head architecture that allows the model to simultaneously predict atmospheric weather and air pollution variables.
    \item \textbf{Frequency-weighted Loss Function:} 
    To address the heavy-tailed distributions of pollutants like PM1, PM2.5, and PM10, we introduce a Frequency-weighted Mean Absolute Error (fMAE).
    % To effectively handle the heavy-tailed distributions of pollutants such as PM1, PM2.5, and PM10, we introduce a Frequency-weighted Mean Absolute Error (fMAE), an adaption of the class balanced loss function described in \cite{cui2019classbalancedlossbasedeffective}.
    \item \textbf{Regional Adaptation:} 
    % Recognizing that the MENA region undergoes some of the highest PM concentrations globally, we provide better forecasts by enhancing the model's capacity to forecast the high pollution levels in the region.
    Recognizing the MENA region's high PM concentrations, we enhance the model to improve the accuracy of forecasts for severe pollution levels in the region.
    \item \textbf{Combined Dataset:}
    To help the model learn the relationships between atmospheric conditions and pollutant levels, we create a comprehensive dataset by aligning existing weather and chemical datasets across spatial and temporal dimensions. 
    % To support the model in learning the relationships between atmospheric conditions and pollutant levels, we collect a combined and comprehensive dataset by incorporating weather variables with air quality variables.
\end{enumerate}

\section{Related Work}
\label{related_work}
% In the recent years, many domains have been integrated with machine learning methods and forecasting air pollutions is one of them. Though physics based technies have been present for the said task, commonly are WRF-Chem, CMAQ. And due to complex nature of air pollution variables, it is very hard to model with physics based models, inducing high uncertaintis and low predcition accuracy. With more effective principles and characteristics, data driven machine learning models are more effective. Due to non linear features in air pollution variables, data driven models are popular choice among the available options.
% Physics based models are grounded with the principles of atmospheric chemistry and physics. Capturing the weather patterns along with air pollutions has been notable things to make air pollution forecasting more effectuve but it is challenging to set accurate initial conditions and along with that computational expense is high for high resolution and complex environments.

In recent years, the integration of machine learning methods into various scientific domains has gained significant attention, with air pollution forecasting a notable example. 
Traditionally, physics-based models like the WRF-Chem \citep{ojha2020widespread} and CMAQ \citep{zhang2012source} model have been employed to predict air pollution levels. 
These models are grounded in the fundamental principles of atmospheric chemistry and physics to simulate complex interactions within the atmosphere.
However, the highly nonlinear and complex nature of air pollution variables poses substantial challenges for these physics-based models. 
Modeling air pollution's complexity often leads to high uncertainties and reduced prediction accuracy \citep{hao2020spatiotemporal, li2019air}. 
Additionally, running these models at high resolutions in complex environments demands significant computational resources, which can be challenging for real-time forecasting.

In contrast, data-driven machine learning models \citep{yu2022deep, cai2023forecasting, bodnar2024aurora} have emerged as a more effective alternative for air pollution forecasting. These models excel at capturing nonlinear relationships and patterns within large datasets, allowing them to handle the complexities of air pollution variables more adeptly. Machine learning approaches can provide more accurate and efficient predictions without requiring detailed physical simulations by learning directly from the data. 
% Consequently, they have become a popular choice aiming to improve air quality forecasts. 
In \cite{yu2022deep}, a deep ensemble-based approach is introduced for estimating daily PM2.5 concentrations. This framework leverages machine learning base models, such as XGBoost, which are used to train meta-models in the second stage, with an optimization algorithm applied in the third stage.
In \cite{cai2023forecasting}, authors proposed a framework to enhance the prediction of hourly PM2.5 concentrations. Their method involves breaking down complex data into simpler components, each representing different frequency levels. These components are then modeled using a combination of autoregressive and CNN-based methods to capture patterns in data, to improve the accuracy of the predictions.

\begin{table*}[t]
\caption{A list of all the weather and air quality variables present in our dataset. Furthermore, for variables that contain data at different pressure levels, we collect 7 of them.}
\label{tab:variables}
\adjustbox{max width=\linewidth}{
\begin{tabular}{@{}llll@{}}
\toprule
& Variable (short name)                                    & Description                                                & Pressure Levels       \\ \midrule
\multirow{9}{*}{\rotatebox[origin=c]{90}{Weather Variables}} &
geopotential (z)                              & Varies with the height of a pressure level                 & 7 levels     \\&
temperature (t)                               & Temperature                                                & 7 levels     \\&
specific humidity (q)                         & Mixing ratio of water vapor                                & 7 levels     \\&
relative humidity (r)                         & Humidity relative to saturation                            & 7 levels     \\&
u component of wind (u)                       & Wind in longitude direction                                & 7 levels     \\&
v component of wind (v)                       & Wind in latitude direction                                 & 7 levels     \\&
2m temperature (t2m)                          & Temperature at 2m height above surface                     & Single level \\&
10m u component of wind (u10)                 & Wind in longitude direction at 10m height                  & Single level \\&
10m v component of wind (v10)                 & Wind in latitude direction at 10m height                   & Single level \\ \midrule
\multirow{12}{*}{\rotatebox[origin=c]{90}{Air Quality Variables}} &
carbon monoxide (co)                          & Carbon monoxide concentrations                             & 7 levels     \\&
ozone (go3)                                   & Ozone concentrations                                       & 7 levels     \\&
Nitrogen monoxide (no)                        & Nitrogen monoxide concentrations                           & 7 levels     \\&
Nitrogen dioxide (no2)                        & Nitrogen dioxide concentrations                            & 7 levels     \\&
Sulphur dioxide (so2)                         & Sulphur dioxide concentrations                             & 7 levels     \\&
Particulate matter d \textless 1 µm (pm1)     & Particulate matter with diameter less than 1 µm            & Single level \\&
Particulate matter d \textless 10 µm (pm10)   & Particulate matter with diameter less than 10 µm           & Single level \\&
Particulate matter d \textless 2.5 µm (pm2.5) & Particulate matter with diameter less than 2.5 µm          & Single level \\&
Total column carbon monoxide (tcco)           & Total amount overall levels & Single level \\&
Total column nitrogen monoxide (tc\_no)       & Total amount overall levels & Single level \\&
Total column nitrogen dioxide (tcno2)         & Total amount overall levels & Single level \\&
Total column ozone (gtco3)                    & Total amount overall levels & Single level \\ \bottomrule
\end{tabular}
}
\end{table*}

Recent advancements in neural network architectures have significantly improved the integration of air quality variables with weather variables in forecasting models. 
For instance, the Aurora model \citep{bodnar2024aurora} primarily trains on weather data and then forecasts air pollution levels as a downstream task.
Similarly, ClimaX \citep{nguyen2023climaxfoundationmodelweather} is an open-source weather model that leverages the vision transformer architecture \citep{dosovitskiy2021imageworth16x16words}. 
Trained on large-scale datasets, it serves as a foundational model by employing a pretext task focused on predicting future time steps randomly sampled within a specified range. 
In its downstream applications, ClimaX handles a variety of tasks across different spatial and temporal scales, including regional weather forecasting. Due to its adaptable and efficient design, we have selected ClimaX to demonstrate our proposed approach. 
A recent study \citep{munir2024efficient}, has explored enhancing ClimaX for MENA weather forecasting using parameter-efficient fine-tuning like LoRA. This underscores the potential of adapting foundational models to meet the challenges of specific regions.
However, it's important to note that ClimaX currently operates at a lower spatial resolution compared to models like Aurora \citep{bodnar2024aurora} and CAMS \citep{CAMS23}. In contrast to large-scale foundational models, existing work on PM forecasting typically employs fewer variables, limiting itself to smaller capacity models or statistical approaches \citep{CabelloTorres2022, Masood2023-mc}. A notable exception is the work of \cite{Sarafian2023-mx}, which uses a transformer-based model to forecast PM10 concentrations using several weather variables, demonstrating their importance in the process. However, most studies in this field, while valuable, often focus on predicting a single PM concentration variable and utilize only a subset of available predictors. Our work aims to address these limitations by adopting a more comprehensive, multi-variable approach.

% In AirCast, we integrate air quality variables directly with weather variables. This integration allows us to overcome the limitations of traditional physics-based models, which may not fully capture complex interactions and data-driven methods that consider only one type of data. 
% By combining both air quality and weather data, our approach is better equipped to model the complex environmental dynamics. This is particularly significant for regions like the MENA, where complex environmental factors demand models capable of effectively handling a diverse set of variables.
In AirCast, we combine air quality and weather data to better capture complex environmental dynamics, addressing the limitations of models that rely on one type of data. This approach is important for regions like the MENA, where diverse environmental factors require models capable of handling intricate interactions effectively.

\begin{figure*}[t]
  \centering
    \includegraphics[width=\textwidth]{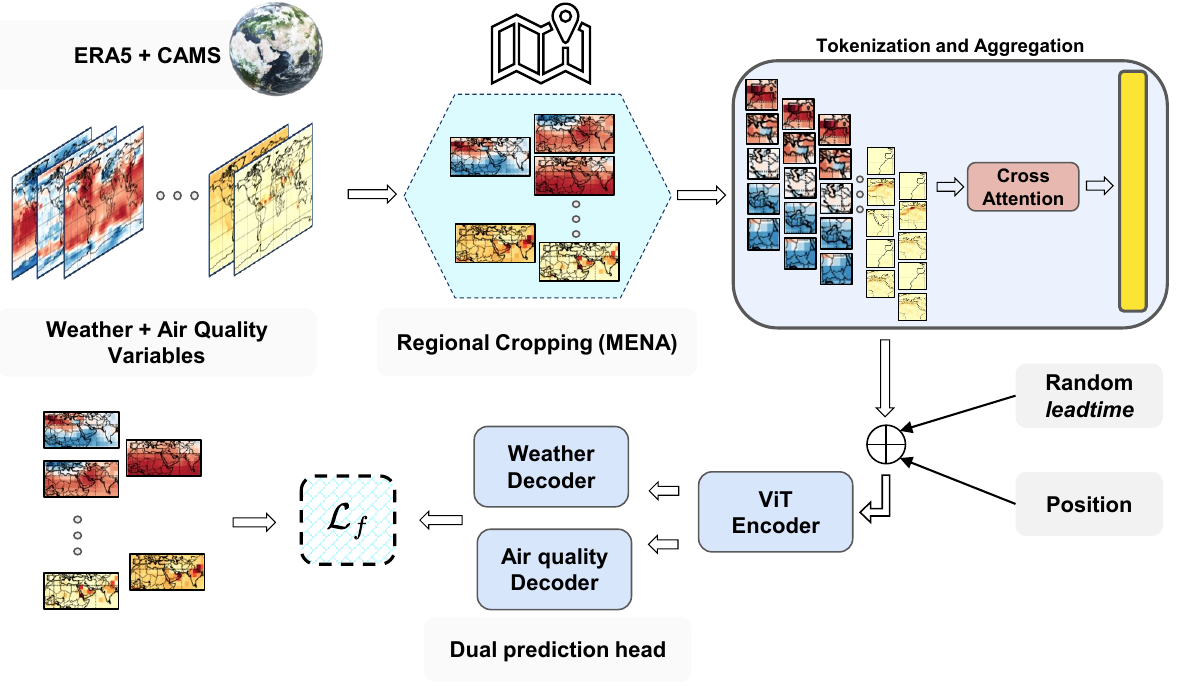}
  \caption{This illustrates the architecture of the \textbf{AirCast} model, an extension of~\cite{nguyen2023climaxfoundationmodelweather}. The model integrates weather data from the ERA5 dataset and air quality data from the CAMS EAC4 dataset. The model is trained using regional data from the MENA region. The input variables are tokenized and aggregated, with a Vision Transformer (ViT) encoder, processing the combined weather and air quality inputs. 
  A dual decoder head is employed, with one predicting weather variables and the other forecasting air quality variables. The predictions are compared with the ground truth at a certain lead time using the Frequency-Weighted MAE loss function. }
  % \vspace{-0.9em}
  \label{fig:method}
\end{figure*}

\section{Aligned Dataset}

We aim to create an aligned dataset with multiple weather and air quality variables (see Table \ref{tab:variables}) to better capture the complex factors influencing PM concentrations.
% This comprehensive dataset is important for meticulously capturing the complex relations that govern air pollution.

\textbf{Weather Variables.} The weather data is sourced from the ERA5 archive \citep{Rasp_2020}, providing hourly data from 1979 to 2018. 
Due to its large size, the dataset has been regridded to resolutions of $5.625^o$ ($32 \times 64$ pixels),
and $1.40525^o$ ($128 \times 256$ pixels). Furthermore, to temporally align with the chemical pollutant variables described next, we only choose the years from 2003 to 2018. 
% This facilitates the model to accurately learn the relationships between weather patterns and pollutant levels.
% We only choose the $5.625^o$ resolution for our experiments to balance data granularity and computational feasibility. 
For our experiments, we focus on the $5.625^o$ resolution to balance data granularity and computational efficiency.

\textbf{Air Quality Variables.} The air quality data is collected from the Copernicus Atmosphere Monitor Service (CAMS) data archive. 
We utilize the ECMWF Atmospheric Composition Reanalysis 4 (EAC4) data catalog, which combines an atmospheric model with real-world observations to create a comprehensive global dataset comprising various air quality variables. 
The data originally comes in a $0.75^o$ resolution and three-hourly intervals. 
For consistency with weather data \citep{Rasp_2020}, we regrid these to $5.625^o$ resolution. 
Furthermore, to align the air quality variables temporally with the weather variables, we interpolate the data to be hourly instead of three hours. 
Following WHO guidelines \citep{WHO2021}, we included additional variables known to affect PM concentrations.
% As suggested by the guidelines created by \cite{WHO2021}, we also collect all other variables that could affect the PM concentrations. 
The full list of air quality variables is shown in Table \ref{tab:variables}.
 
The combined dataset contains both surface and pressure-level variables. For pressure-level data, we selected seven pressure levels: 50, 250, 500, 600, 700, 850, and 925 hecto-Pascals (hPa). These levels were chosen to represent a broad range of atmospheric dynamics, from near-surface to higher altitudes.
% The full dataset consists of both surface variables (variables whose values are recorded at the surface) and pressure variables (variables whose values are recorded at various pressure levels). In the case of the pressure variables, we only choose the 50, 250, 500, 600, 700, 850 and 925 hPa. 
The unit hPa is typically used to represent different vertical levels in the atmosphere, with a pressure of approximately 1000 hPa at sea level, decreasing as altitude increases.
% Such granularity is vital for understanding the complex structures of PM atmospheric dynamics. 

\begin{figure}[t]
    \centering
    \includegraphics[width=0.9\linewidth]{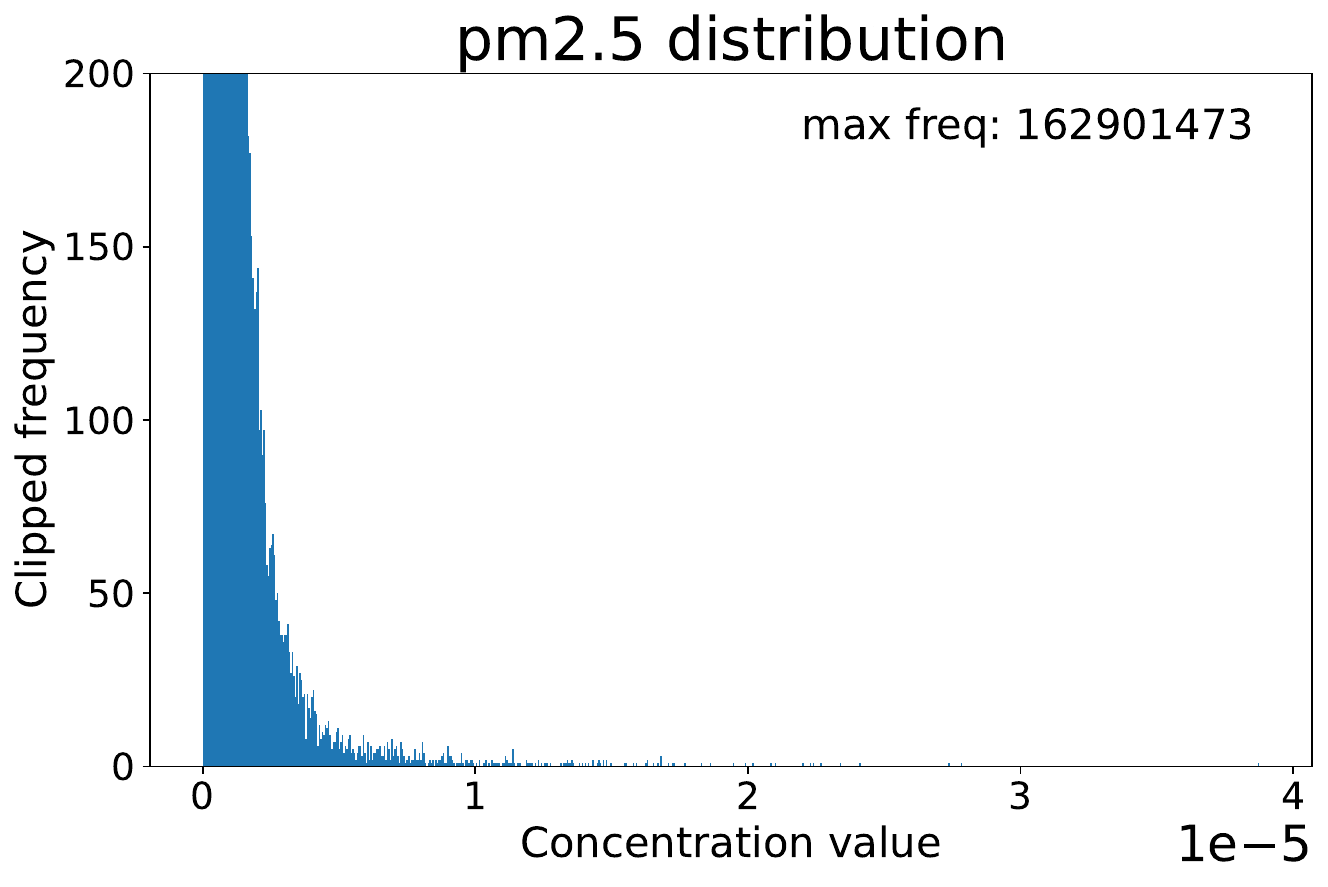}
    \caption{\textbf{Skewed distribution of PM2.5.} The y-axis corresponds to the frequency clipped at 200 (the maximum frequency is shown in each figure).
    }
    \label{fig:distribution_pm2p5}
\end{figure}

\textbf{Distribution Skew.} 
% A key point to note is that several chemical variables have a heavy-tailed distribution (the PM variables are shown in Figure \ref{fig:distribution_plots}). 
Many air quality variables show heavy-tailed distributions, notably for PM concentrations, as shown in Figure~\ref{fig:distribution_pm2p5}.
% The distribution of other variables can be found in the Appendix. For example, in the case of PM2.5, the high concentration values indicate high air pollution. 
% From the distribution, it can be seen that the amount of pixels with such a high PM concentration is quite low. 
% This particularly makes the task of forecasting quite challenging. 
% This phenomenon means that high pollution levels are relatively infrequent, but can be significant when they occur. 
% For example, high PM2.5 concentrations, while infrequent, are important indicators of severe air quality problems.
This phenomenon indicates that while high pollution levels, including PM1, PM2.5, and PM10, are rare, they have a significant impact when they occur. These elevated concentrations, though infrequent, are critical indicators of severe air quality issues.
% Precise modeling and forecasting these rare events is a challenge, adding complexity to the dataset. 

\section{AirCast}
% \subsection{AirCast}
In this section, we describe our approach AirCast (Figure~\ref{fig:method}),  for multi-variable air pollution forecasting.
Inspired by ClimaX~\citep{nguyen2023climaxfoundationmodelweather}, we use a Vision Transformer ViT \citep{dosovitskiy2021imageworth16x16words}  as the backbone.
% \cite{nguyen2023climaxfoundationmodelweather} pre trained ClimaX using various climate and weather datasets. 
The model was pre-trained using a variety of climate and weather datasets, each with a varying number of variables \cite{nguyen2023climaxfoundationmodelweather}.
Similar to ClimaX, the variable tokenization module was utilized to standardize the input and a variable aggregation module was employed to handle the large sequence of input variables during training, thereby reducing the sequence length and enhancing computational efficiency.

\textbf{Variable Tokenization} is a process that converts each input variable separately into a sequence of patches. 
Specifically, each input variable $V$ of size $H \times W$ is tokenized in a sequence of size $H/p \times W/p$, where $p$ denotes the size of the patch. 
The input patches are then passed through an embedding layer, resulting in a sequence of dimensions $V \times H/p \times W/p \times D$, where $D$ denotes the embedding dimension.
% By processing each variable independently, the model can capture variable specific patterns before integrating them.

\textbf{Variable Aggregation} follows the variable tokenization, using a cross-attention mechanism to aggregate information from multiple variables at the same spatial location.
% variables for the same spatial location. 
% This helps in reducing the information from all variables which results in a sequence length of shape  ($H/p \times W/p$). 
This process effectively reduces the sequence length to ($H/p \times W/p$) while retaining essential information from all input variables.
This aggregation not only optimizes computational efficiency but also improves the model’s ability to understand the relations between weather and air quality variables.

Since we aim to use weather and air quality variables, an additional prediction head is added. 
% For both heads the number of output variables are same as the number of input variables. 
Both prediction heads output the same number of variables as the input.
% Through some experiments, we find that this setup results in the best performance. 
The loss is calculated independently for each set of variables: weather loss is computed for the weather variables from the weather head, and likewise for the air quality head.
By decoupling the learning processes, the model alleviates potential negative transfer between tasks, which is usually a challenge in multi-task learning frameworks.
Experimental evidence shows that this configuration yields the best performance.
% One thing to note is that while our output variables for both heads and the input variables are the same, the loss is only computed on the variables associated with that head i.e. we compute the weather loss on the weather variables from the weather head and similarly for the chemical head. 

\subsection{Regional Setup}
We target a specific region instead of forecasting the entire globe. 
% Note that our model can easily be adapted globally, but to better evaluate its forecasting capabilities, we decided to choose MENA region with very high levels of PM concentrations thereby indicating a high pollution level. 
% % We chose the MENA region (Middle East and North Africa) for all the experiments. 
% The MENA region has some of the highest PM concentrations globally \citep{Li2022, Nissenbaum2023}, often exceeding the WHO guidelines significantly by a large margin.
While our model can be adapted globally, we focus on the MENA region to evaluate its forecasting capabilities due to the region’s high PM concentrations. 
The MENA region consistently records some of the highest PM levels worldwide \citep{Li2022, Nissenbaum2023}, frequently exceeding WHO guidelines.
We expect that by choosing such a region, we can focus our model capability on forecasting the higher PM concentrations.   

\subsection{Normalization}
% \textbf{Normalization.} 
% Before the input data is passed through the model, the weather variables undergo standard normalization, while the air quality variables undergo standard normalization followed by a scaled log transformation (shown in equation~\ref{eq:4}. 
% This log transformation emphasizes smaller values, which are often dominated by larger ones.
% This process helps in stabilizing the training and better captures the variability of smaller concentration values in air quality variables.
% Note that the frequency computation is performed on the data after applying normalization and log transformation. 
Prior to the model training, weather variables are normalized, while air quality variables undergo normalization followed by a scaled log transformation (shown in equation~\ref{eq:4}). 
The log transformation highlights smaller values often overshadowed by larger ones, stabilizing training and capturing the variability of low air quality concentrations more effectively.
This log transformation is inverted during validation and test time. 

\begin{equation}
    \label{eq:4}
    x = \frac{\log(\max(x, 10^{-4})) - \log(10^{-4})}{\log(10^{-4})}
\end{equation}

% \textbf{Normalization.} Before the input data is passed through the model, the weather variables undergo standard normalization, while the chemical variables undergo standard normalization followed by a scaled log transformation (shown in equation~\ref{eq:4}. 
% This log transformation emphasizes smaller values, which are often dominated by larger ones.
% This process helps in stabilizing the training and better captures the variability of smaller concentration values in chemical variables.
% Note that the frequency computation is performed on the data after applying normalization and log transformation. 
% This log transformation is inverted during validation and test time. 

% \begin{equation}
%     \label{eq:4}
%     x = \frac{\log(\max(x, 10^{-4})) - \log(10^{-4})}{\log(10^{-4})}
% \end{equation}

\subsection{Randomized Lead Time}
While our experiments focus on forecasting the variables, 24 hours from the input time (lead time), we find that randomizing the lead time during training results in improving the model performance. 
We believe this acts as an extra augmentation technique that may serve as a regularization, by exposing the model to various forecasting horizons. 
For each training sample, the lead time is randomly chosen from 6, 12, and 24 hours intervals.
% Hence during training, for sample, we choose randomly between $6hr$, $12hr$, and $24hr$ lead times. 
% We only validate and test on a $24hr$ lead time. 
For validation and testing, however, only a 24-hour lead time is used to maintain consistency.
% By training the model on multiple lead times, it becomes more robust and can capture temporal traits more effectively (see Appendix for results).

\subsection{Frequency-Weighted Mean Absolute Error} 
Many air quality variables, including PM1, PM2.5, and PM10, exhibit a heavy-tailed distribution (as illustrated in ~Figure \ref{fig:distribution_pm2p5} for PM2.5).
To address this skewness, we propose a Frequency-weighted Mean Absolute Error (fMAE) function motivated by class-balancing approaches \citep{cui2019classbalancedlossbasedeffective}.
% To address this skewness, we propose a Frequency-weighted Mean Absolute Error (fMAE) function aimed at addressing the problem of class imbalance.
% We pre-compute the frequency of the values for each chemical variable based on an optimal set of bin widths derived using the Freedman Diaconis Estimator. Based on the frequency, we apply the weight as shown in 
The frequency of values for each air quality variable in the training data is pre-computed using optimal bin widths specified by the Freedman-Diaconis Estimator \citep{Freedman1981}. 
Based on this frequency, a weight is assigned according to Equation \ref{eq:1},

\begin{equation}
\label{eq:1}
    W_{freq} = \begin{cases}
        0\ , \ freq = 0 \\
        \frac{1-\beta}{1-\beta^{freq}}, \ otherwise \\
    \end{cases}
\end{equation}

% The intuition is that if the frequency of a bin is high, then we want to give that a lower weight and vice versa. 
where $\beta$ is a hyperparameter that is used to define the frequency weighting term. $\beta \rightarrow 0$ signifies equal weighting while $\beta \rightarrow 1$ signifies inverse frequency weighting. 
Based on experimentation, we found that setting $\beta$ to 0.8 resulted in the best performance.
The core idea behind this weighting scheme is to provide greater emphasis on rare events and reduce the impact of frequently occurring events.
% This allows pixel values that are underrepresented to have a higher weight when computing the loss. 
% This weighting ensures that the model pays more attention to underrepresented scenarios, such as extreme pollution events, thereby enhancing its predictive accuracy for these critical cases.
% We choose the value 0.2 to prevent a weight of 0 when the frequency is 1. 

Additionally, following the methodology used in WeatherBench \citep{Rasp_2020}, we further employ the latitude weight along with the frequency-weighted loss. 
The latitude weight is defined in Equation \ref{eq:2} and is used to account for the varying sizes of grid cells due to the Earth's spherical shape.
% to account for the spherical nature of the Earth to give a different weight to smaller grid cells than larger grid cells.

\begin{equation}
\label{eq:2}
W_{lat}^i = \frac{\cos(lat(i))}{\frac{1}{H}\sum_{i^` = 1}^H\cos(lat(i^`))}, 
\end{equation}

Where $lat(i)$ is the latitude of the $i^{th}$ row in the grid, $H$ corresponds to the height of the image and $W_{lat}$ is the latitude weight for each $i$. 
% Since weather variables do not have a heavy-tailed distribution, we just use the latitude weighting factor for the weather loss. 
The overall loss is described in Equation \ref{eq:3}

% \begin{equation}
%     \label{eq:3}
%     % \small
%     \mathcal{L}_f = (W_{lat} \times MAE_{weather}) + \\ 
%     (W_{freq} \times W_{lat} \times MAE_{chemical})
% \end{equation}

\begin{equation}
    \label{eq:3}
    \begin{aligned}
        \mathcal{L}_f &= (W_{lat} \times MAE_{weather})~+ \\
        &\quad (W_{freq} \times W_{lat} \times MAE_{chemical})
    \end{aligned}
\end{equation}

% where the second term in the summation is what we call the frequency-weighted MAE (fMAE) and this term also includes the latitude weight.
This dual-weighting approach helps the model sufficiently capture the weather and air quality variables' spatial and distributional variations.
By including latitude weights, the model accounts for the variations in grid cell areas at different latitudes, which is crucial for global-scale modeling.
The frequency weights, on the other hand, address the imbalance in the distribution of air quality concentrations, enhancing the model's ability to predict rare events.

With these improvements, AirCast provides an adaptive framework for multi-variable air pollution forecasting, leveraging weather and air quality data to improve accuracy in high-pollution regions.

% By incorporating these improvements, AirCast demonstrates a robust and adaptive framework for multi-variable air pollution forecasting.
% It effectively leverages weather and air quality data to enhance predictive accuracy, particularly for regions with severe pollution challenges.

\section{Experimental Setup}

\subsection{Implementation details}
% When training AirCast on the new air pollution data, we first initialize the network with the pre-trained weights from the original ClimaX paper. 
For training AirCast on the new combined dataset, the network is initialized with pre-trained weights from \cite{nguyen2023climaxfoundationmodelweather}.
We use a learning rate of $5\times10^{-4}$, a batch size of $32$ and a seed of $42$ when training the model. The original shape of the input is $32 \times 64$ at $5.625^o$ resolution and after applying the regional cropping, the input is $8 \times 14$. 
% We train the model for $100$ epochs with early stopping. The full model with all the variables was trained on 4 A100 GPUs and roughly took 6 hours. 
The model is trained for 100 epochs with early stopping criteria to prevent overfitting. The training is conducted on four A100 GPUs, taking approximately four hours for the model with all variables. 
The dataset was temporally partitioned to create train, validation, and test splits. Specifically, data from 2003 to 2015 was allocated to the training set, while 2016 data constituted the validation set. The test set comprised data from 2017 and 2018.

\subsection{Baselines}
To evaluate AirCast's performance, we compare it against two established models: a persistence baseline and the CAMS global atmospheric composition forecast. 
Notably, to the best of our knowledge, only one other study (Aurora, \cite{bodnar2024aurora}) has attempted to forecast all three PM variables using a single model. 
However, their repository does not provide access to fine-tuned models for air pollution forecasting, precluding a direct comparison.

\textbf{Persistence Baseline:} 
% This straightforward model assumes that the forecast for the next 24 hours will be identical to the current input. Despite its simplicity, it is a useful benchmark for assessing more complex models.
This baseline model predicts that the forecast over the next 24 hours will remain unchanged from the current input. While simple, it is a valuable benchmark for evaluating the performance of more complex forecasting models.

\textbf{CAMS Global Forecasts:} The CAMS global atmospheric composition forecast is a comprehensive data catalog that provides twice-daily forecasts for various lead times. The forecast is generated by using a physics based atmospheric model that learns the complex patterns of several concentrations.

We conduct our baseline evaluations exclusively on data from 2017. We standardize the input time to 00:00 and use a 24-hour lead time for all forecasts.

\section{Results}
\begin{table*}[!ht]
\caption{AirCast Ablations: Various ablations that result in the best performing setting. The reported metric is Root Mean Squared Error - RMSE (lower is better). The unit for all PM concentration variables is $\mu g m^{-3}$. (\textit{3 PM} correspond to PM2.5, PM10, PM1. The ablations in all tables use a lead time of \textit{24 hrs} during testing. Air Quality (\textit{AQ}) corresponds to the full list of air quality variables as shown in Table~\ref{tab:variables}, which includes the 3 PM variables. \textit{Surface} corresponds to the near-surface pressure level of multi-level variables (high pressure).  $\neg$ is used when we consider the low pressure levels of multi-level variable. For each table, the initial setting (to compare against) is defined in \colorbox{baselinecolor}{gray}. The best setting is defined in \colorbox{bestsettingcolor}{yellow}. }
\label{results}
\begin{subtable}{.49\textwidth}
    \small
    \centering
    \captionsetup{width=.69\textwidth}
    \caption{\textbf{Impact of the fMAE loss}. We only consider the 3 PM variables as input and output. }
    \label{results_a}
    \begin{tabular}{@{}lc@{}}
    \toprule
    Method                & \begin{tabular}[l]{@{}c@{}} RMSE ($\mu g m^{-3}$) \\ PM2.5\ /\ PM10\ /\ PM1 \end{tabular}\\ \midrule
    \baseline{without fMAE} & \baseline{10.05\ /\ 15.34\ /\ 7.38} \\ 
    \ with fMAE           & \textbf{9.63}\ /\ \textbf{14.78}\ /\ \textbf{7.17} \\ \bottomrule
    \end{tabular}
\end{subtable} %
% \hspace{1cm}
\hfill
\begin{subtable}{.49\textwidth}
    \small
    \centering
    \captionsetup{width=.69\textwidth}
    \caption{\textbf{Adding additional variables}}
    \label{results_b}
    \begin{tabular}{@{}lc@{}}
    \toprule
    Variables                & \begin{tabular}[l]{@{}c@{}} RMSE ($\mu g m^{-3}$) \\ PM2.5\ /\ PM10\ /\ PM1 \end{tabular}\\ \midrule
    \baseline{3 PM} & \baseline{9.63\ /\ 14.78\ /\ \textbf{7.17}} \\
    Weather + 3 PM           & 9.94\ /\ 14.72\ /\ 7.69    \\
    AQ &  9.80\ /\ 14.96\ /\ 7.41\\ 
    Weather + AQ & \textbf{9.45}\ /\ \textbf{14.15}\ /\ 7.24    \\
    \bottomrule
    \end{tabular}
\end{subtable}
\begin{subtable}{.49\textwidth}
    \small
    \centering
    % \vspace{1em}
    \captionsetup{width=.69\textwidth}
    \caption{\textbf{Considering near surface variables.}}
    \label{results_c}
    \begin{tabular}{@{}lc@{}}
    \toprule
    Variables                & \begin{tabular}[l]{@{}c@{}} RMSE ($\mu g m^{-3}$) \\ PM2.5\ /\ PM10\ /\ PM1 \end{tabular}\\ \midrule
    \baseline{Weather + AQ} & \baseline{9.45\ /\ 14.15\ /\ 7.24} \\
    Surface Weather + AQ  & 9.24\ /\ 13.81\ /\ 7.14 \\
    Weather + Surface AQ  & 9.65\ /\ 14.08\ /\ 7.60 \\ 
    \bestsetting{Surface Weather + Surface AQ}  & \bestsetting{\textbf{8.82}\ /\ \textbf{13.27}\ /\ \textbf{6.65}} \\ 
    $\neg$ Surface Weather + $\neg$ Surface AQ          & 9.40\ /\ 14.09\ /\ 7.19 \\
    \bottomrule
    \end{tabular}
\end{subtable}
\hfill
\begin{subtable}{.49\textwidth}
    \small
    \centering
    % \vspace{1em}
    \captionsetup{width=.69\textwidth}
    \caption{\textbf{Baseline comparison} }
    \label{results_d}
    \begin{tabular}{@{}lc@{}}
    \toprule
    Method              & \begin{tabular}[l]{@{}c@{}} RMSE ($\mu g m^{-3}$) \\ PM2.5\ /\ PM10\ /\ PM1 \end{tabular}\\ \midrule
    Persistence Baseline & 13.12\ /\ 20.00\ /\ 9.88 \\
    CAMS forecasts  & 22.16\ /\ 32.01\ /\ 18.21 \\
    \bestsetting{Surface Weather + Surface AQ}  & \bestsetting{8.82\ /\ 13.27\ /\ 6.65} \\ 
    \bottomrule
    \end{tabular}
\end{subtable}
% \begin{subtable}{.49\textwidth}
%     \small
%     \centering
%     \hspace{0.5em}
%     \captionsetup{width=.69\textwidth}
%     \caption{\textbf{Considering non surface variables}}
%     \label{results_c}
%     \begin{tabular}{@{}lc@{}}
%     \toprule
%     Variables                & \begin{tabular}[l]{@{}c@{}} RMSE ($\mu g m^{-3}$) \\ PM2.5\ /\ PM10\ /\ PM1 \end{tabular}\\ \midrule
%     $\neg$ Surface Weather + $\neg$ Surface AQ          & 9.40\ /\ 14.09\ /\ 7.19    \\
%     Surface Weather + Surface AQ  & \textbf{8.82}\ /\ \textbf{13.27}\ /\ \textbf{6.65} \\ \bottomrule
%     \end{tabular}
% \end{subtable} %

\end{table*}

\begin{figure*}[!]
    \centering
      \begin{subfigure}{0.48\textwidth}
        \label{vis_a}
        \includegraphics[width=\textwidth]{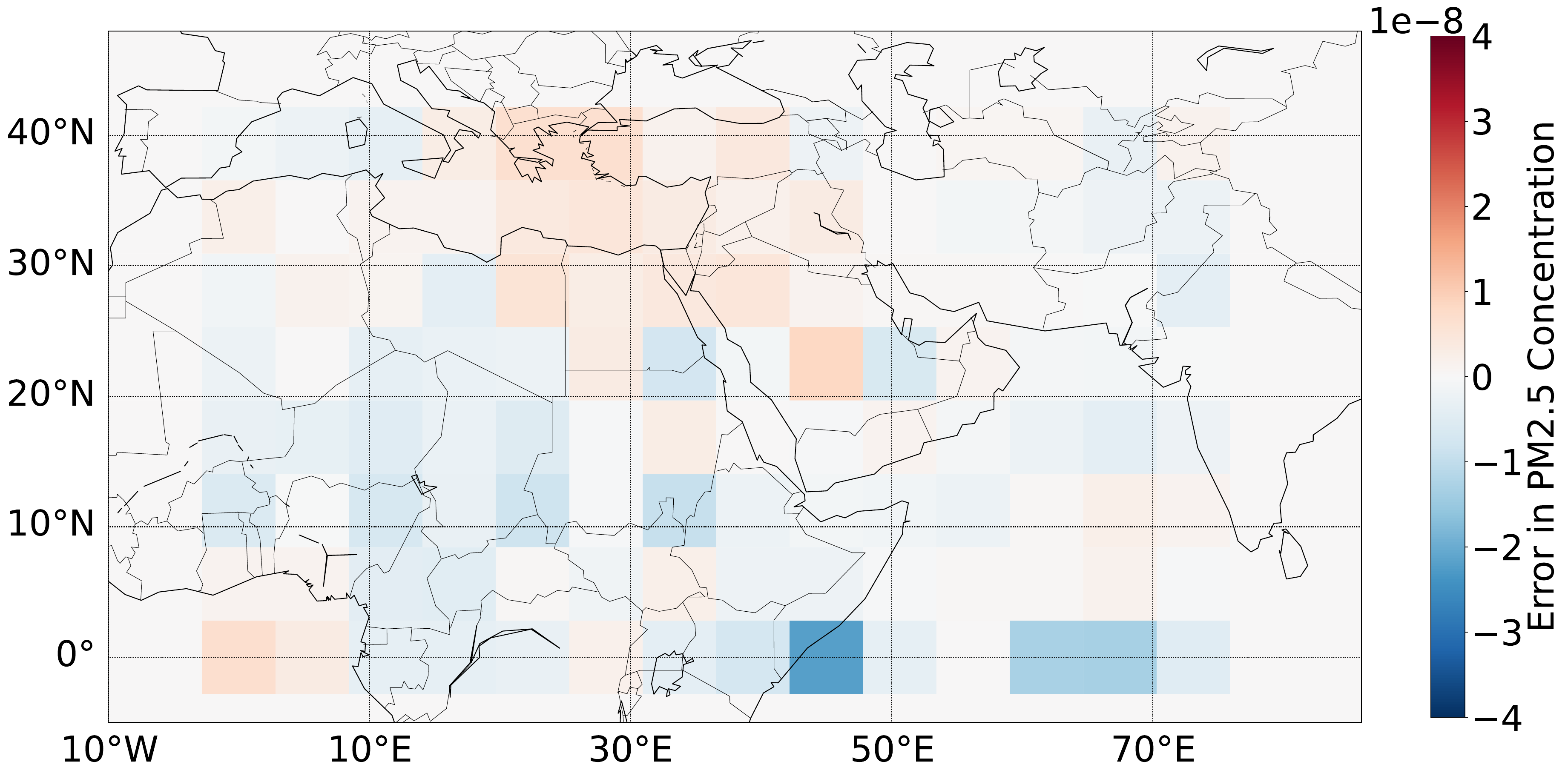}
      \end{subfigure}
      \begin{subfigure}{0.48\textwidth}
        \label{vis_b}
        \includegraphics[width=\textwidth]{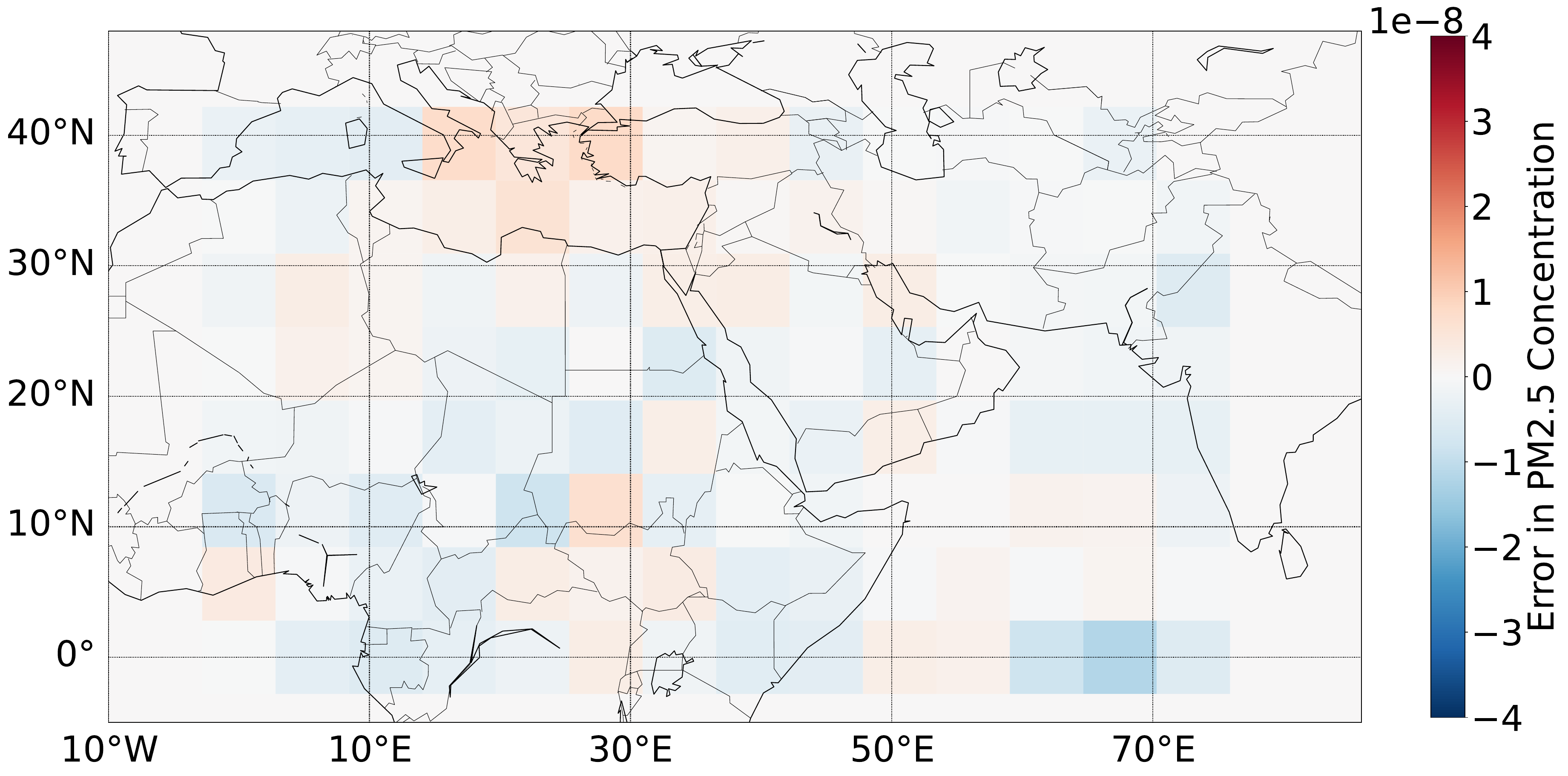}
      \end{subfigure}
      % \begin{subfigure}{0.32\textwidth}
      %   \includegraphics[width=\textwidth]{figures/vis_error_3.pdf}
      % \end{subfigure}

    \caption{Sample error plots for PM2.5 forecasting (prediction - ground truth). The unit is $kg m^{-3}$. The first and second plots are without and with the proposed fMAE loss function respectively. }
    \label{fig:vis}

\end{figure*}

% In this section, we present and analyze the results of our air pollution forecasting experiments, focusing on PM2.5, PM10, and PM1 concentrations. Our investigation systematically explores the impact of different input variables, data transformations, and model configurations on forecasting performance, with the goal of developing a robust and reliable air pollution forecasting approach. While our primary focus is on PM concentrations, we have conducted comprehensive analyses for all output variables, with additional details available in the Appendix.

In this section, we analyze the results of the air pollution forecasting experiments, focusing on PM1, PM2.5, and PM10 concentrations. 
We systematically examine the effects of various input variables, data transformations, and model configurations on forecasting performance. 

% (input and output variables are the same). These results are shown in the Appendix. 

% \textbf{Adding more weather variables improves the forecasting of PM.}
% \textbf{How does the fMAE loss affect PM forecasting?}
\textbf{Impact of fMAE Loss on PM Forecasting: }
Considering the heavy-tailed distribution of several air quality variables, an experiment was conducted with and without the proposed fMAE loss. 
% It is important to note that we only consider the 3 PM concentration variables as both input and output. 
Results from Table \ref{results_a} indicate that using fMAE loss led to an improvement of forecasting RMSE of 4.18\%, 3.65\% and 2.85\% for PM2.5, PM10, PM1 respectively. 
For this particular experiment, we report numbers using only the three PM concentration variables as both input and output.
% Furthermore, the visualization when using the fMAE loss (Figure \ref{fig:vis}) results in a slightly better forecasting model for higher PM concentrations (denoted by the lighter blue areas in the second plot). 
Furthermore, visualizations using the fMAE loss (Figure \ref{fig:vis}) indicate a slightly better forecasting model for higher PM concentrations, as denoted by the lighter blue areas in the second plot.
% \begin{table}[!]
%     \small
%     \centering
%     \label{results_a}
%     \begin{tabular}{@{}lc@{}}
%     \toprule
%     Variables                & \begin{tabular}[l]{@{}c@{}} RMSE ($\mu g m^{-3}$) \\ PM2.5\ /\ PM10\ /\ PM1 \end{tabular}\\ \midrule
%     PM           & 10.36\ /\ 15.81\ /\ 7.81    \\
%     PM + \\ Weather & \textbf{9.70}\ /\ \textbf{14.30}\ /\ \textbf{7.53} \\ \bottomrule
%     \end{tabular}
% \end{table} %

% \textbf{What is the effect of incorporating additional weather variables and air quality variables on PM forecasting?} 

\textbf{Impact of Weather and Air Quality Inputs on PM Forecasting:}
% Table \ref{results_b} contains results from an experiment conducted to analyse the effect of adding more weather and air quality variables in addition to the three PM concentration variables. 
% Note that air quality variables defined in \ref{tab:variables} also contain the three PM concentration variables. 
% Adding all variables led to a improvement in forecasting RMSE of 1.87\%, 4.26\% for PM2.5 and PM10 respectively, and a slight degradation for PM1. While we expected the forecasting RMSE for PM1 to also be improved, the outcome for PM2.5 and PM10 is consistent with findings in the existing literature and confirms the significant correlation between weather variables and PM concentrations~\citep{CabelloTorres2022, Yang2017}. 
% Furthermore the WHO guidelines \citep{WHO2021} suggest the correlation between several other air quality variables and the PM concentrations.
Table \ref{results_b} presents results from an experiment analyzing the impact of incorporating additional weather and air quality variables.
% alongside the three PM concentration variables. 
Notably, air quality variables (in Table \ref{tab:variables}) include these PM concentrations as well, and we denote them as AQ.
Incorporating all variables improved the forecasting RMSE by 1.87\% for PM2.5 and 4.26\% for PM10 but slightly degraded performance for PM1. While an improvement in PM1 forecasting was anticipated, the results for PM2.5 and PM10 align with existing literature, highlighting the strong correlation between weather variables and PM concentrations ~\citep{CabelloTorres2022, Yang2017}. 
Additionally, WHO guidelines \citep{WHO2021} underscore the links between various air quality variables and PM concentrations.

% \textbf{What is the effect of selectively choosing only near surface level variables?} 
\textbf{Effect of Selecting Near-Surface Variables: }
% As demonstrated by \cite{Li2017, Sarafian2023-mx}, near surface level variables are very crucial in forecasting PM concentrations. To verify this, we experiment (Table \ref{results_c}) by only considering the near surface level variables among the multi-level variables, and we consider all the single-level variables directly. 
As shown by \cite{Li2017, Sarafian2023-mx}, near-surface-level variables are crucial for forecasting PM concentrations. 
To verify this, we conduct experiments (Table \ref{results_c}) using only near-surface-level variables from multi-level data while directly including all single-level variables.
The resulting forecasting RMSE is improved by 6.67\%, 6.22\%, and 8.15\% for PM2.5, PM10, PM1 respectively, when considering surface-level weather and air quality variables. 
% To also verify that the model is not performing well only due to fewer variables, 
We conduct another experiment where we consider the low-pressure level variables (represented by $\neg$). 
The results confirm that selecting variables strongly correlated with PM concentrations significantly enhances forecasting accuracy.
% It is evident that choosing the right variables that are more correlated with PM concentrations results in an improved forecasting model.

% \textbf{How does the best setting compare with other baselines?} 
\textbf{Baselines Comparison with Our Best Model: } 
% As mentioned above, there exists no single model that forecasts all PM concentration variables and we cannot directly compare to Aurora due to non availability of the fine-tuned model. 
% Hence, we select the persistence baseline which is a simple yet effective baseline to compare against, and the CAMS global forecasts. 
% From table \ref{results_d}, our model performs significantly better than the persistence baseline and the CAMS global forecasts. 
As mentioned earlier, no single model currently forecasts all PM concentration variables, and a direct comparison with Aurora is also not possible due to the unavailability of its fine-tuned model for air quality forecasting. 
Therefore, we use the persistence baseline, a simple yet effective benchmark, and CAMS global forecasts for comparison (see Table \ref{results_d}. 
% As shown in Table \ref{results_d}, our model significantly outperforms both the persistence baseline and CAMS global forecasts.
\begin{figure*}[t]
    \centering
    \begin{subfigure}{0.48\linewidth}
        \includegraphics[width=\textwidth]{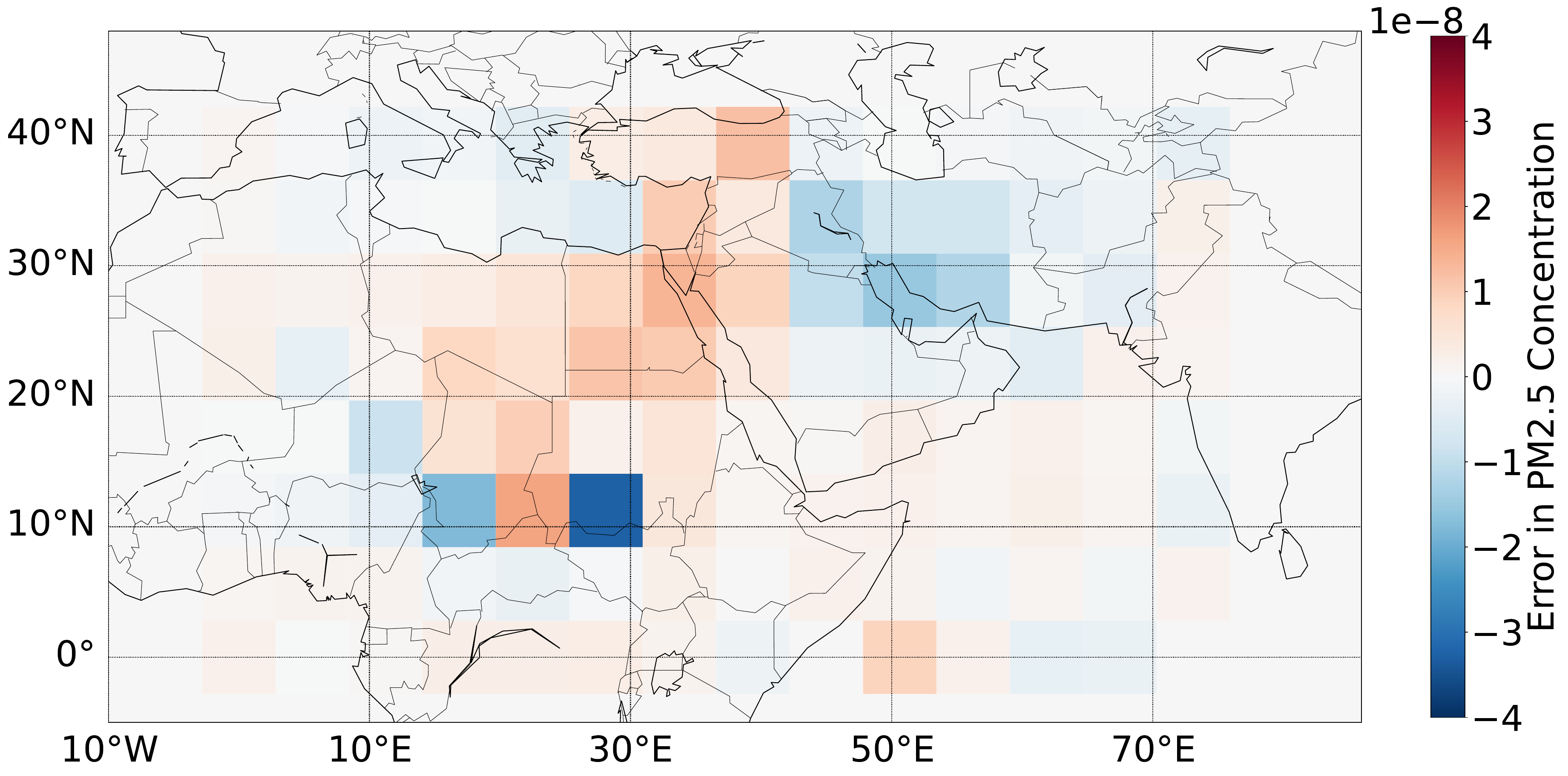}
        \caption{Saudi Arabia - October 29, 2017 (CAMS Forecasts)}
        % \label{fig:extreme_2}
    \end{subfigure}
    \centering
    \begin{subfigure}{0.48\linewidth}
        \includegraphics[width=\textwidth]{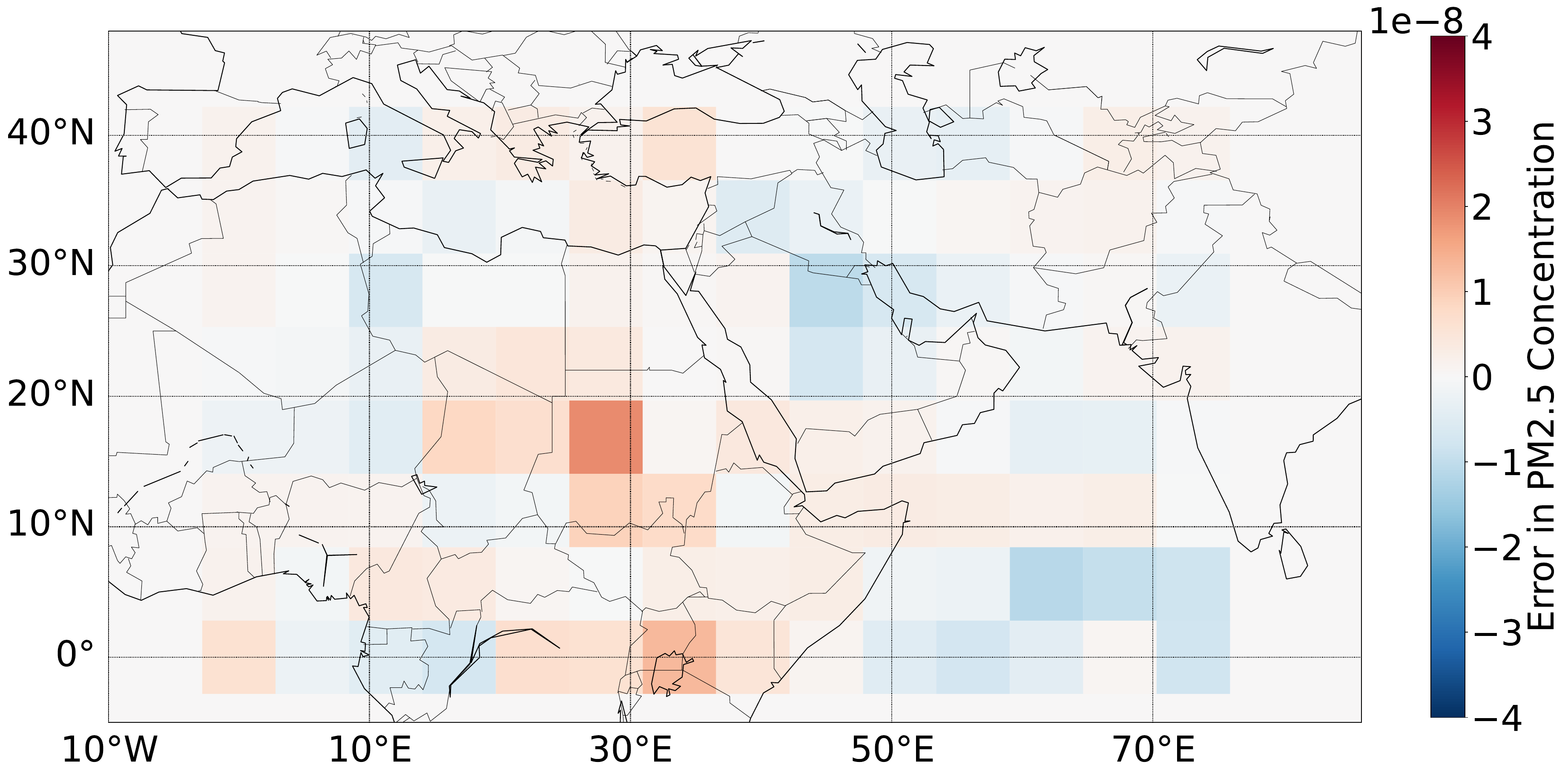}
        \caption{Saudi Arabia - October 29, 2017 (Aircast)}
        % \label{fig:extreme_1}
    \end{subfigure}

    \centering
    \begin{subfigure}{0.48\linewidth}
        \includegraphics[width=\textwidth]{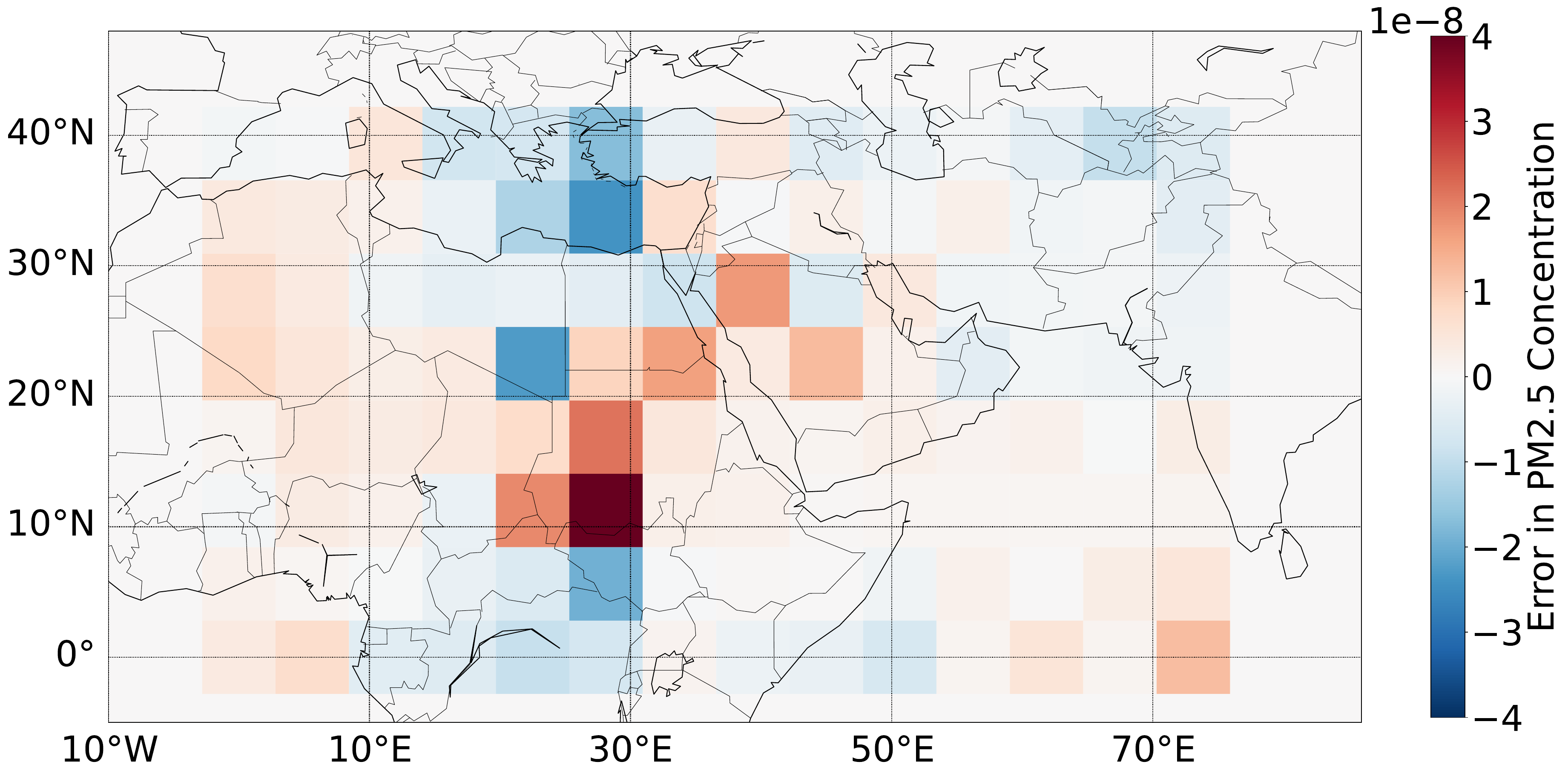}
        \caption{Kuwait - October 31, 2017 (CAMS forecasts)}
        % \label{fig:extreme_2}
        \end{subfigure}
    \centering
    \begin{subfigure}{0.48\linewidth}
        \includegraphics[width=\textwidth]{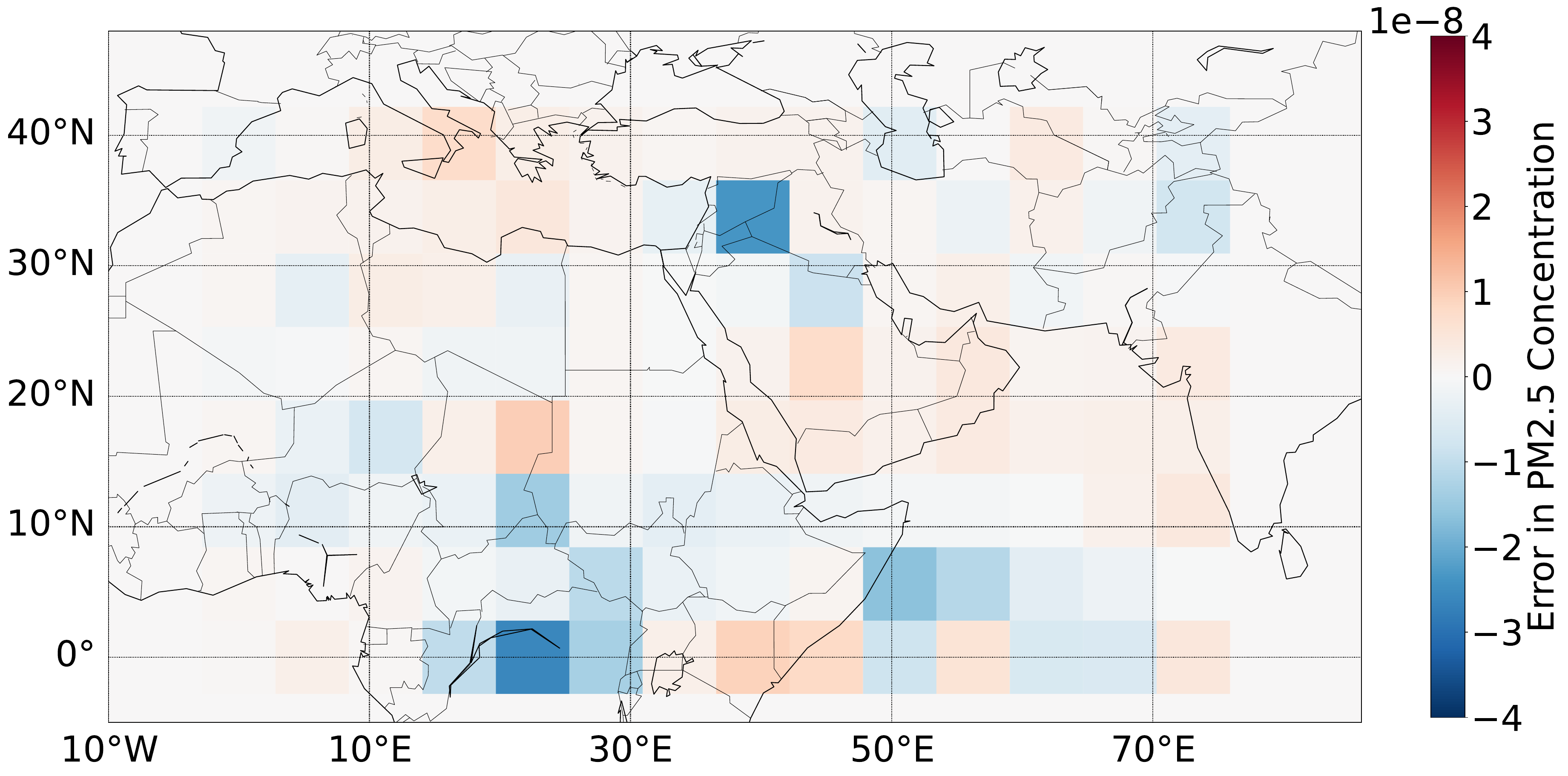}
        \caption{Kuwait - October 31, 2017 (Aircast)}
        % \label{fig:extreme_1}
    \end{subfigure}
    \caption{Extreme case visualizations of PM2.5 concentrations (Predictions - Ground Truth) for CAMS global forecasts and Aircast.}
    \label{fig:extreme}
\end{figure*}

\textbf{Extreme Event Forecasting:} To test our models' capability in forecasting extreme events, we selected two dust storm events: one in Kuwait on October 31, 2017, and another in Saudi Arabia on October 29, 2017. While there is room for improvement, our model demonstrated the ability to detect these events. This can be observed in Figure \ref{fig:extreme}, where light blue or white colors in the affected regions closely correspond to the actual dust storm occurrences.

\textbf{Randomized lead time:} Following \cite{nguyen2023scalingtransformerneuralnetworks}, we investigate the benefit of randomizing the lead time during training and validation. Results from table \ref{app_1} indicate that randomizing lead time improves the PM forecasting performance. We believe this acts as an
extra augmentation technique and allows the model to learn from various forecasting horizons. 
\begin{table}[t]
    \small
    \centering
    % \captionsetup{width=.69\textwidth}
    \caption{\textbf{Randomized lead time.} An ablation to test the performance with and without the randomized lead time (6, 12, 24 hrs) during training and validation. At test time, the lead time is fixed to 24hrs.}
    \label{app_1}
    \begin{tabular}{@{}lc@{}}
    \toprule
    Method                & \begin{tabular}[l]{@{}c@{}} RMSE ($\mu g m^{-3}$) \\ PM2.5\ /\ PM10\ /\ PM1 \end{tabular}\\ \midrule
    Without & 9.04\ /\ 13.61\ /\ 6.80 \\
    With  & \textbf{8.82}\ /\ \textbf{13.27}\ /\ \textbf{6.65} \\ 
    \bottomrule
    \end{tabular}
\end{table}

\begin{table}[t]
    \small
    \centering
    % \captionsetup{width=.69\textwidth}
    \caption{\textbf{Varying lead time.} Considering only near surface weather and air quality variables.}
    \vspace{1em}
    \label{app_2}
    \begin{tabular}{@{}lc@{}}
    \toprule
    Lead time                & \begin{tabular}[l]{@{}c@{}} RMSE ($\mu g m^{-3}$) \\ PM2.5\ /\ PM10\ /\ PM1 \end{tabular}\\ \midrule
    6 hrs & 6.30\ /\ 9.26\ /\ 5.04 \\
    12 hrs  & 7.60\ /\ 11.2\ /\ 6.03 \\
    24 hrs  & 8.82\ /\ 13.27\ /\ 6.65 \\ 
    48 hrs  & 11.50\ /\ 18.01\ /\ 7.97 \\ 
    \bottomrule
    \end{tabular}
\end{table}

\textbf{Varying lead times:} To test the models forecasting performance at various temporal times, we run an additional experiment by varying the lead time. Results from Table \ref{app_2} indicate that there is an inverse relation between the lead time and forecasting RMSE. This suggests that our models forecasting ability is more robust in near-term. 

\section{Conclusion}
% Prior work in air pollution forecasting has generally involved statistical-based models, and traditional machine learning approaches or have resorted to using fewer variables. 
Previous efforts in air pollution forecasting have primarily relied on statistical models, traditional machine learning approaches, or limited variable sets.
% In this work, we present a multi-variable approach to forecasting several chemical variables (mainly focusing on PM concentrations). 
In this work, we propose a multi-variable approach with a particular focus on forecasting PM concentrations. 
% We create a dataset comprising chemical pollutant data and weather data that have been spatially and temporally aligned with each other. We further present AirCast, a ViT-based forecasting model that makes use of all these different variables.
We develop a spatially and temporally aligned dataset that integrates chemical pollutant and weather data. 
Building on this, we introduce AirCast, a Vision Transformer (ViT)-based forecasting model that leverages these diverse variables.
Our results demonstrate that incorporating weather and air quality variables significantly enhances PM forecasting accuracy. 
Notably, near surface-level variables emerge as the most impactful in driving the synergy between weather and air quality data.
% Additionally, selecting specific features informed by prior studies further boosts model performance. 
To address the heavy-tailed distribution of chemical variables, we introduce a Frequency-weighted Mean Absolute Error (fMAE) loss function, which effectively captures rare the high pollution events. 
Finally, we will make our code and dataset fully open-source to facilitate future advancements in air pollution forecasting.
% We first show how adding weather variables can improve PM forecasting performance. We also show that choosing specific features based on previous studies can further improve the performance of the model. 
% Finally, to better learn the heavy-tailed distribution observed in the chemical variables, we introduce a new Frequency-based MAE loss. We also make the code and data fully open source to enable further research in this domain. 

\section*{Impact Statement}

Accurate air pollution forecasting is crucial for protecting public health and informing environmental policy decisions. From a public health perspective, reliable forecasts enable individuals with respiratory conditions or other sensitivities to take precautionary measures, reducing their exposure to harmful pollutants. This approach can potentially lead to decreased healthcare costs and an improved quality of life for vulnerable populations.
Furthermore, precise forecasting empowers government and healthcare systems to better prepare for and respond to air pollution events. By anticipating periods of poor air quality, authorities can implement timely interventions, such as issuing public health advisories or temporarily restricting high-emission activities. While the method described in this paper is only for a relatively short lead time, this sets the road for future work that can improve forecasts for longer periods of time.

% In the unusual situation where you want a paper to appear in the
% references without citing it in the main text, use \nocite
% \nocite{langley00}

% \clearpage
% \newpage
\bibliography{example_paper}
\bibliographystyle{icml2025}

%%%%%%%%%%%%%%%%%%%%%%%%%%%%%%%%%%%%%%%%%%%%%%%%%%%%%%%%%%%%%%%%%%%%%%%%%%%%%%%
%%%%%%%%%%%%%%%%%%%%%%%%%%%%%%%%%%%%%%%%%%%%%%%%%%%%%%%%%%%%%%%%%%%%%%%%%%%%%%%
% APPENDIX
%%%%%%%%%%%%%%%%%%%%%%%%%%%%%%%%%%%%%%%%%%%%%%%%%%%%%%%%%%%%%%%%%%%%%%%%%%%%%%%
%%%%%%%%%%%%%%%%%%%%%%%%%%%%%%%%%%%%%%%%%%%%%%%%%%%%%%%%%%%%%%%%%%%%%%%%%%%%%%%

\newpage
\clearpage
\appendix
% \onecolumn
\section{Appendix}

\subsection{Geographic Generalization:} 
\begin{figure}[h!]
    \centering
    \begin{subfigure}{0.9\linewidth}
        \includegraphics[width=\textwidth]{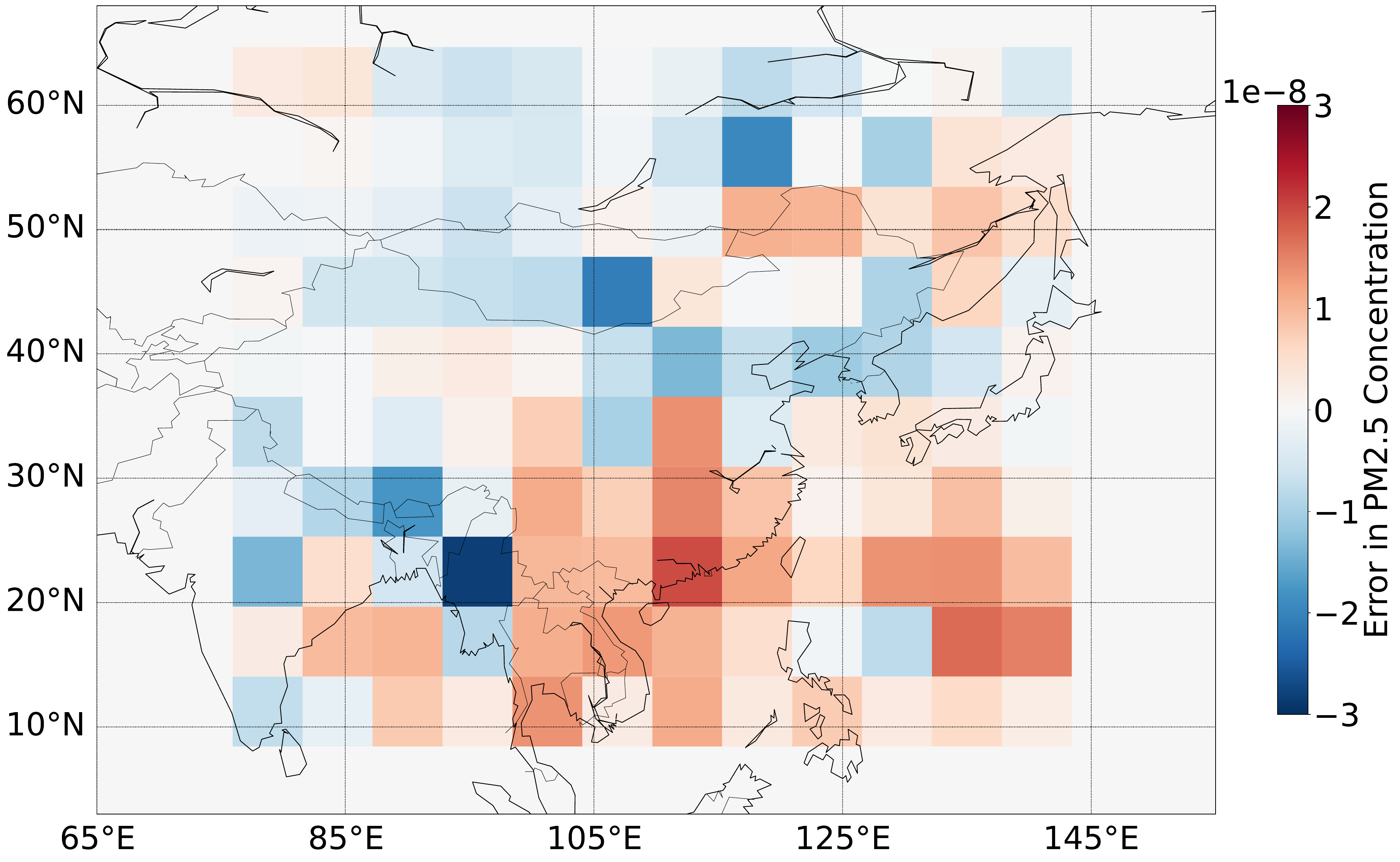}
        \caption{East Asia}
        % \label{fig:extreme_1}
    \end{subfigure}
    \centering
    \begin{subfigure}{0.9\linewidth}
        \includegraphics[width=\textwidth]{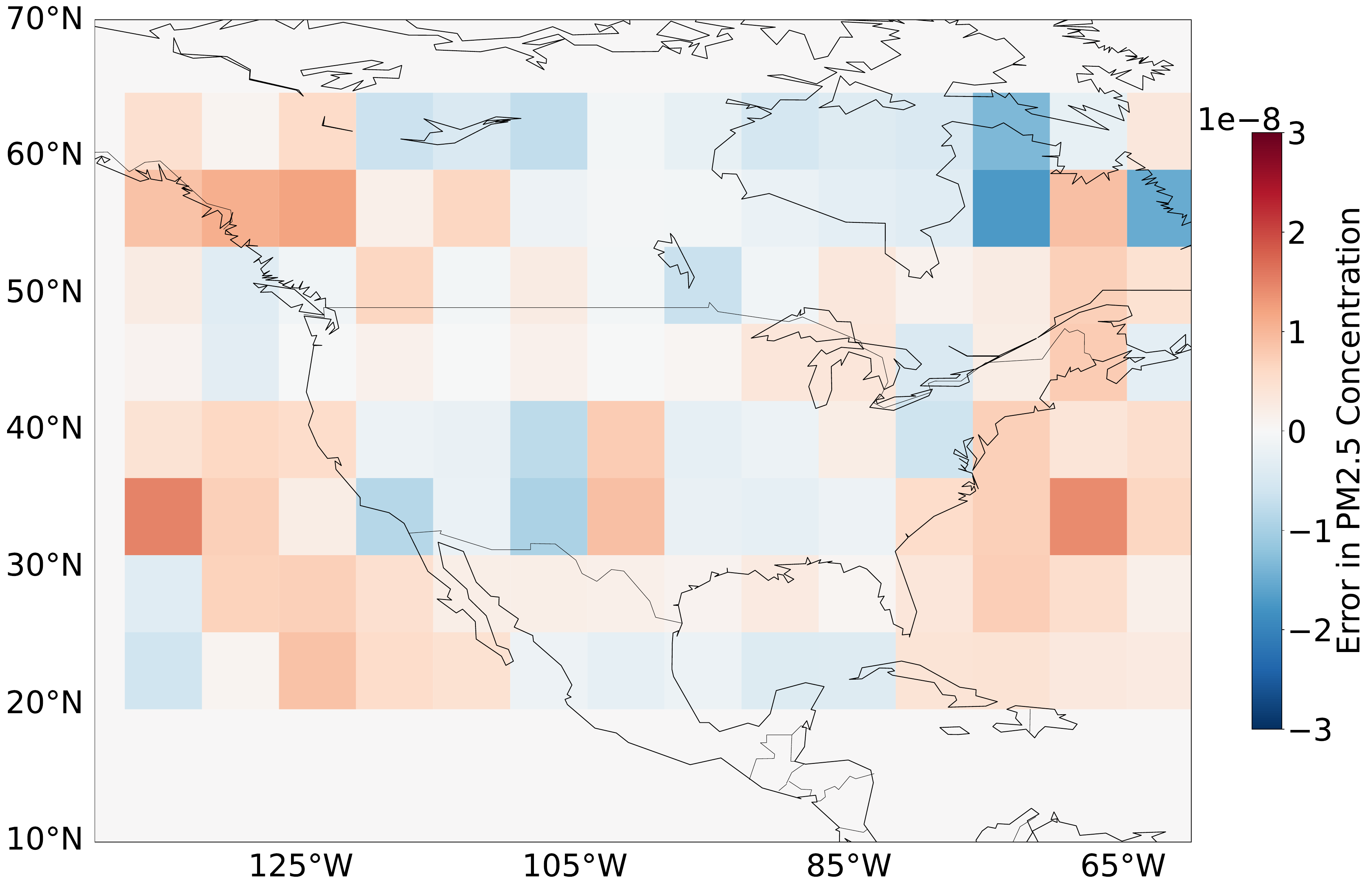}
        \caption{North America}
        % \label{fig:extreme_2}
        \end{subfigure}
    \caption{Extreme case visualizations of PM2.5 concentrations (Predictions - Ground Truth) for Aircast and the CAMS global forecasts}
    \label{fig:geo_gen}
\end{figure}

While we only focus the training on the MENA region, we test the geographic generalization ability of our model in East Asia and North America. We find that Aircast slightly over-estimates in some areas (denoted by the red areas in Figure \ref{fig:geo_gen}).

% \begin{table}[h!]
%     \small
%     \centering
%     % \captionsetup{width=.69\textwidth}
%     \caption{Geographic Generalization: Testing our best model in East Asia and North America with a lead time of 24hrs. }
%     \label{geo_gen}
%     \begin{tabular}{@{}lc@{}}
%     \toprule
%     Region                & \begin{tabular}[l]{@{}c@{}} RMSE ($\mu g m^{-3}$) \\ PM2.5\ /\ PM10\ /\ PM1 \end{tabular}\\ \midrule
%     East Asia & 10.89\ /\ 20.61\ /\ 3.74 \\
%     North America  & 9.26\ /\ 18.22\ /\ 1.89 \\ 
%     \bottomrule
%     \end{tabular}
% \end{table}
\begin{table}[h!]
    \small
    \centering
    \caption{Geographic Generalization: Testing our best model in East Asia and North America with a lead time of 24hrs.}
    \label{geo_gen}
    \begin{tabular}{@{}c@{\hspace{1em}}c@{\hspace{1em}}c@{\hspace{1em}}c@{}}
    \hline
    \\[-0.8em]
    \multicolumn{1}{c}{\multirow{2}{*}{Region}} & \multicolumn{3}{c}{RMSE ($\mu g m^{-3}$)} \\ 
    \\[-0.8em]
    \cline{2-4}
    \\[-0.8em]
    & PM2.5 & PM10 & PM1 \\ 
    \hline
    \\[-0.8em]
    East Asia & 10.89 & 20.61 & 3.74 \\
    North America & 9.26 & 18.22 & 1.89 \\ 
    \\[-0.8em]
    \hline
    \end{tabular}
\end{table}

\subsection{Varying Seeds}
All the ablations in the paper were performed when considering a seed of 42. We additionally test our best setting by varying between 5 different seeds, and report the mean and std. This is a common practice in machine learning research to ensure reproducibility.  

\begin{table}[h!]
    \small
    \centering
    \caption{Varying Seeds for our Best Model. We report the mean, and the standard deviation is reported in the brackets.}
    \label{seeds}
    \begin{tabular}{@{}c@{\hspace{1em}}c@{\hspace{1em}}c@{\hspace{1em}}c@{}}
    \hline
    \\[-0.8em]
    \multicolumn{1}{c}{\multirow{2}{*}{Variables}} & \multicolumn{3}{c}{RMSE ($\mu g m^{-3}$)} \\ 
    \\[-0.8em]
    \cline{2-4}
    \\[-0.8em]
    & PM2.5 & PM10 & PM1 \\ 
    \hline
    \\[-0.8em]
    Best Setting & 9.00 (0.11) & 13.61 (0.20) & 6.78 (0.11) \\ 
    \\[-0.8em]
    \hline
    \end{tabular}
\end{table}

\subsection{Distribution plots of the PM concentrations} 
\begin{figure}[h!]
  \centering
  \begin{subfigure}{0.4\textwidth}
    \includegraphics[width=\textwidth]{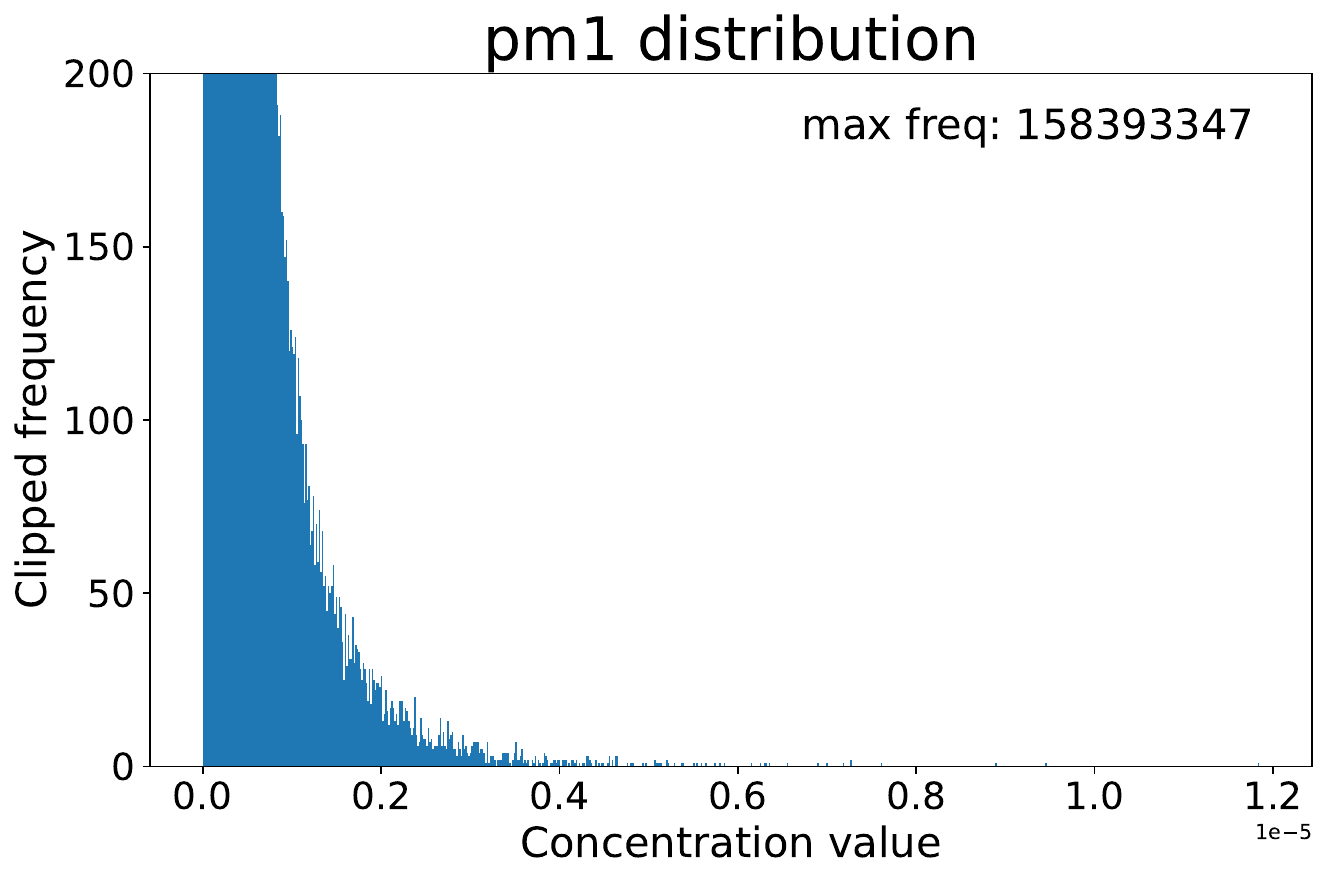}
  \end{subfigure}
  \begin{subfigure}{0.4\textwidth}
    \includegraphics[width=\textwidth]{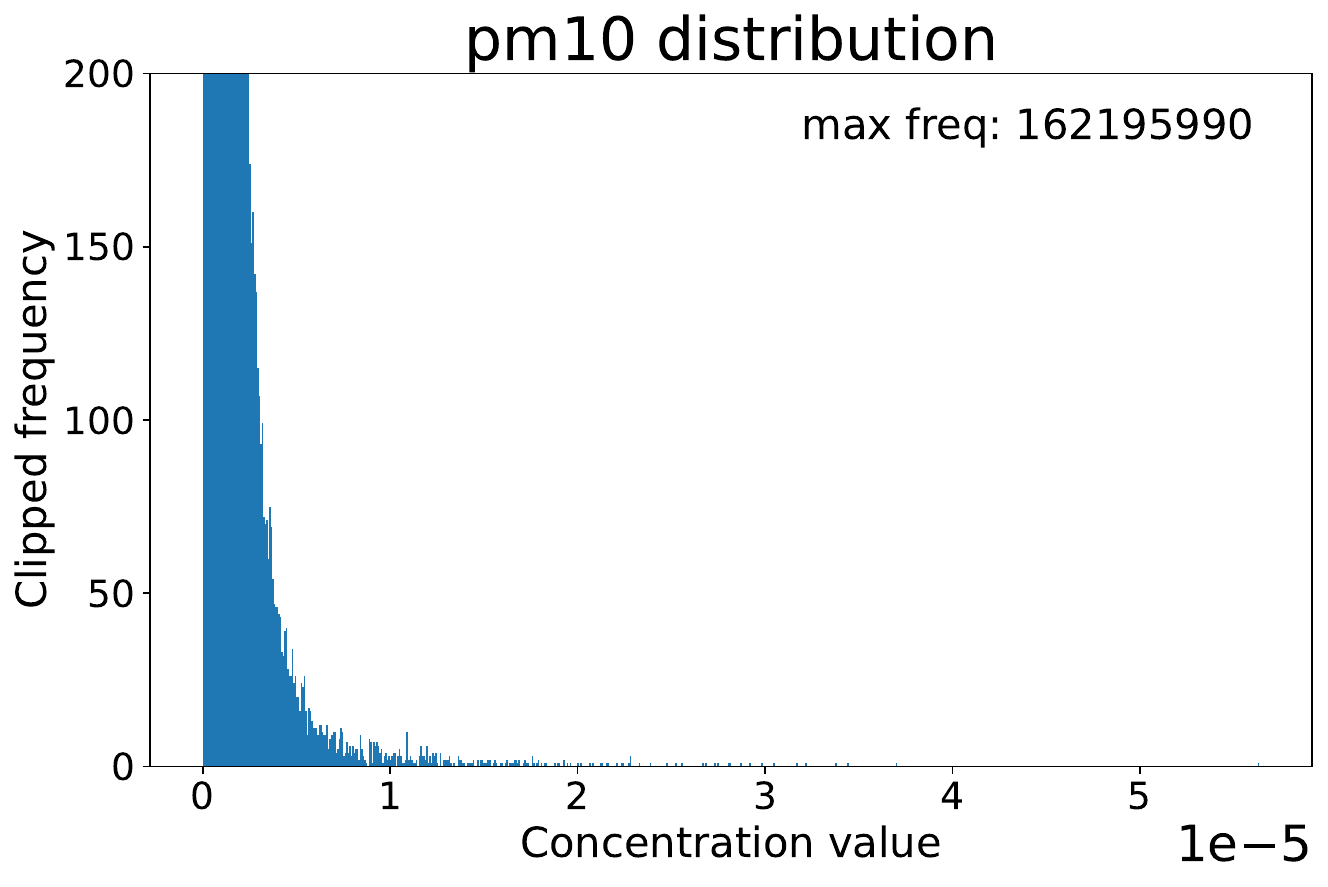}
  \end{subfigure}
  \begin{subfigure}{0.4\textwidth}
    \includegraphics[width=\textwidth]{figures/pm2p5.pdf}
  \end{subfigure}

  \caption{
    \textbf{Skewed distribution of the PM variables.} The x-axis corresponds to the PM variable, and the y-axis corresponds to the frequency clipped at 200 (the maximum frequency is shown in each figure). The clipping is done to visualize the distribution among the low-frequency bins. All the concentration values are in the order of $10^{-5}$.
    }
  \label{fig:app_distribution_plots}
\end{figure}

We further show the distribution plots of PM2.5, PM10 and PM1 in Figure \ref{fig:app_distribution_plots}. Similar to PM2.5, the other 2 concentrations variables PM10 and PM1 also show a long tailed distribution.

%%%%%%%%%%%%%%%%%%%%%%%%%%%%%%%%%%%%%%%%%%%%%%%%%%%%%%%%%%%%%%%%%%%%%%%%%%%%%%%
%%%%%%%%%%%%%%%%%%%%%%%%%%%%%%%%%%%%%%%%%%%%%%%%%%%%%%%%%%%%%%%%%%%%%%%%%%%%%%%

\end{document}